\documentclass[10pt, a4paper]{article}

\usepackage[final]{lrec-coling2024} 

\usepackage{graphicx}
\usepackage{subcaption}
\usepackage{colortbl}
\usepackage{xspace}
\definecolor{Gray}{gray}{0.9}

\usepackage{booktabs} 

\title{Conversations as a Source for Teaching Scientific Concepts at Different Education Levels}

\name{Donya Rooein, Dirk Hovy} 

\address{Bocconi University \\
         Milan, Italy \\
         {donya.rooein, dirk.hovy}@unibocconi.it\\}

\abstract{
Open conversations are one of the most engaging forms of teaching. However, creating those conversations in educational software is a complex endeavor, especially if we want to address the needs of different audiences. While language models hold great promise for educational applications, there are substantial challenges in training them to engage in meaningful and effective conversational teaching, especially when considering the diverse needs of various audiences. No official data sets exist for this task to facilitate the training of language models for conversational teaching, considering the diverse needs of various audiences.
This paper presents a novel source for facilitating conversational teaching of scientific concepts at various difficulty levels (from preschooler to expert), namely dialogues taken from video transcripts.
We analyse this data source in various ways to show that it offers a diverse array of examples that can be used to generate contextually appropriate and natural responses to scientific topics for specific target audiences. It is a freely available valuable resource for training and evaluating conversation models, encompassing organically occurring dialogues. While the raw data is available online, we provide additional metadata for conversational analysis of dialogues at each level in all available videos.
\\ \newline  }

\begin{document}

\maketitleabstract

\section{Introduction}
\label{sec:intro}
Language adaptation in teaching is essential for optimal learning outcomes, especially when addressing learners of varying educational levels. As learners grow and progress from one educational level to another, their cognitive abilities, comprehension levels, and subject familiarity evolve~\cite{national2015transforming}. Tailoring teachers' language to suit the learner's educational level ensures that the content is neither too simplistic nor overwhelmingly complex. By adapting language to a student's academic maturity, educators can create a more inclusive, effective, and engaging learning environment that acknowledges and addresses the unique needs of every learner~\cite{islam2003adapting}. 

The majority of NLP research studies have concentrated on text simplification to enhance comprehension by addressing the syntactic and lexical complexities of a sentence~\cite{siddharthan2002architecture,zhu2010monolingual,alva2020asset}. Limited research has been conducted on sentence \textit{complexification}, which augments the syntactic and lexical intricacy of a provided sentence~\cite{berov2018discourse}. Complexification in learning refers to the situation when a learner might want to see more advanced or idiomatic ways of learning content~\cite{chi2023learning}. The task of paraphrasing a sentence to a specific level of complexity within NLP has been relatively overlooked~\cite{bhagat2013paraphrase,chi2023learning}.


\begin{figure}[ht!]
    \centering
    
    \begin{subfigure}{\columnwidth}
        \centering
        \includegraphics[width=\linewidth]{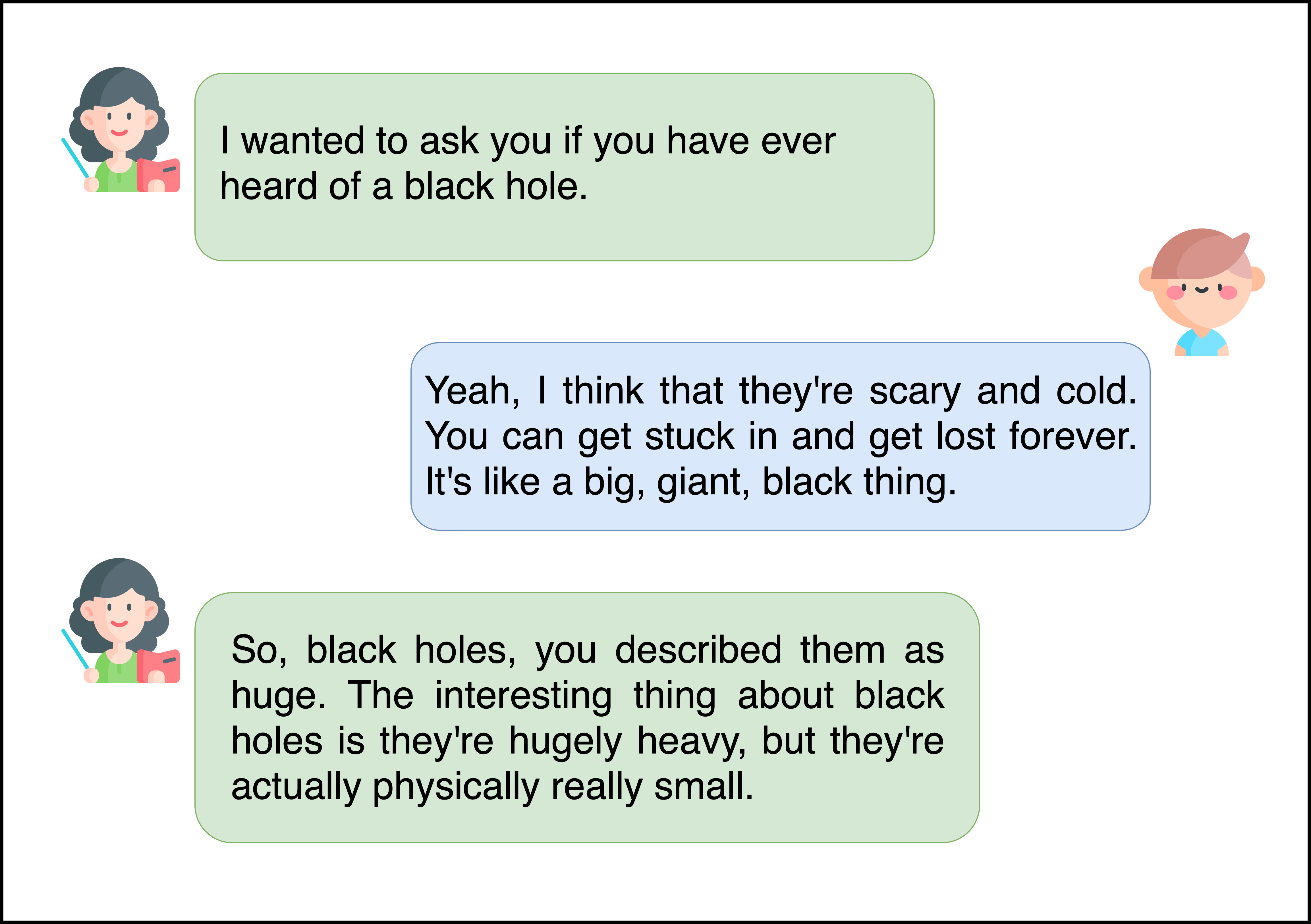}
        \vspace{0.1cm}
    \end{subfigure}

    \begin{subfigure}{\columnwidth}
        \centering
        \includegraphics[width=\linewidth]{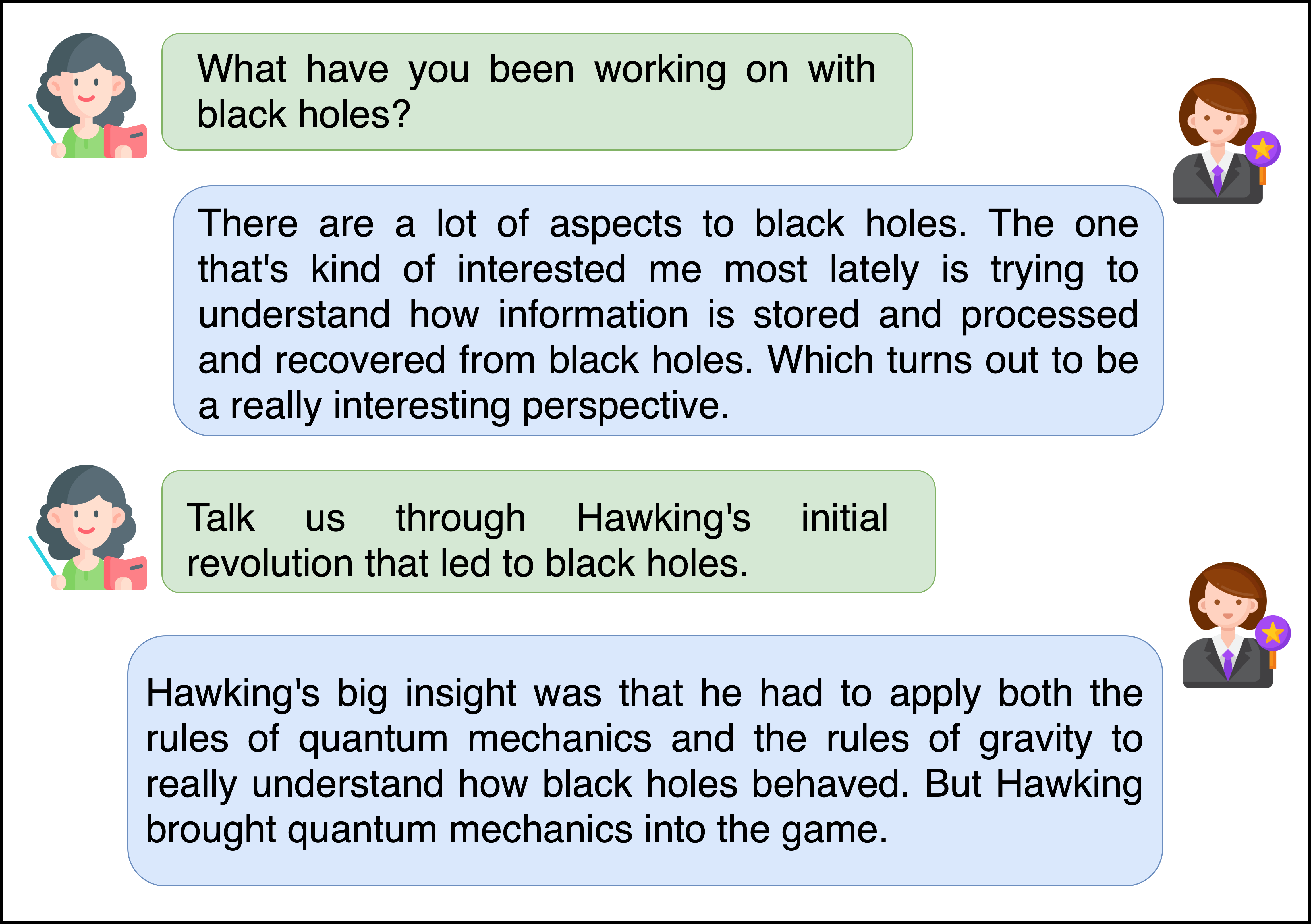}
    \end{subfigure}

    \caption{Examples of conversations of an instructor with a child and expert learners.} 
    \label{fig:conv-example}
\end{figure}



As Large Language Models (LLMs) continue to gain prominence and exhibit remarkable linguistic powers, the educational landscape stands to benefit significantly. The advent of these advanced language models opens the door to developing teaching materials and educational chatbots that are powerful and exceptionally flexible. However, to address this challenge, we need to collect and analyse datasets to adapt language models to serve diverse audience levels. We present a comprehensive study on available sources to facilitate the construction of adaptable educational resources with different complexity levels. We consider this conversational dataset a valuable testing ground to evaluate the adaptability of LLMs to a wide range of audience levels.
Ultimately, our research endeavors to bridge the gap between LLMs' immense promise and their (still limited) real-world applicability in education, thus charting a path toward a more accessible and personalized learning experience for all.
In this paper, we collect and analyse the 5-Levels conversation dataset, an annotated resource to facilitate the advancement of educational dialogue systems capable of adapting to various levels of complexity. It builds on publicly available raw data of educational conversations and adds an analysis and meta-data to make it actionable for LLM research. 
Specifically, the dataset addresses simplifying and complexifying the same learning topic within conversational data to teach scientific concepts across various difficulty levels. It can offer valuable insights into how to address the challenge of developing a language-teaching chatbot capable of customizing the learning experience to suit its audience.

Our contributions are:

\begin{itemize}
\item We collect and analyse the 5-Levels dataset, a high-quality conversational dataset sourced from the WIRED website, featuring interactions between instructors and learners of varying backgrounds in general scientific topics. 

\item We publish the script for collecting video transcriptions to facilitate future research.

\item We quantitatively and qualitatively analyse the interactions between instructors and learners for different levels, providing relevant insights with various statistical metrics.

\end{itemize}

\section{5-Levels Videos}
``5-Levels'' is a video series by WIRED \footnote{\href{https://www.wired.com/video/series/5-levels}{5-Levels videos from Wired, Condé Nast}} to answer the question, ``Can everything be explained to everyone in terms they can understand?''. 
In each video, an expert scientist explains a high-level scientific subject in five different levels of complexity: first to a child, then a teenager, then an undergrad majoring in the same subject, a graduate student, and finally, an expert colleague. Each video segment features an engaging and interactive dialogue among participants. Like personalized one-on-one instruction, the teacher initiates conversations and queries learners on various concepts.

The pairs of dialogue transcription and education level in the 5-level videos offer single-round dialogues between instructors and learners with varying subject knowledge levels. The publicly available transcriptions of these dialogues serve as a rich source for text simplification and language adaptation when conveying scientific concepts to diverse audiences. These transcriptions are self-contained among each complexity level, implying that no prior background or contextual information is required to grasp the core of the conversation. All videos are multi-modal conversations, but the transcription alone suffices to comprehend the dialogue, even without its visual contents. These videos were published from 2017 to 2023 as 25 episodes\footnote{Twenty-three videos were recorded as in-person teaching discussions, two as remote video discussions.} covering various general science topics spanning physics, mathematics, astronomy, machine learning, and cryptography.
The list of participants' gender is available in Table~\ref{tab:demography}. The instructors in these videos are between 38 and 67 years old, with a minimum qualification of a Master of Science degree, and the specific ages of the participants are not explicitly provided. Based on the transcriptions, the age of child participants ranged from 5 to 10 years old, teens were within the 11 to 18 age bracket, college students were 19 to 22 years, and graduate students were typically 21 years or older.

\begin{table}
    \centering
    \begin{tabularx}{\columnwidth}{|l|X|X|X|X|X|X|}
        \hline
        & Ins. & child & teen & coll. & grad. & exp. \\
        \hline
        female & 10 & 17 & 18 & 13 & 12 & 7 \\
        \hline
        male & 15 & 8 & 7 & 12 & 13 & 18 \\
        \hline
    \end{tabularx}
    \caption{Gender distribution of participants in the videos.}
    \label{tab:demography}
\end{table}

\section{Transcriptions}
The transcriptions of the videos are available on the WIRED website. We obtain these transcriptions through a semi-automated method.\footnote{We obtained permission from WIRED to publish our findings on their data.} 
First, we parse all the HTML pages for the videos and extract the transcriptions from them. Then, we manually segment and refine the transcriptions for learners at different levels and separate the turns of instructors and learners. 
Each video consists of seven segments. The initial and final segments are introductions and conclusions, where the instructor addresses the topic. The remaining segments feature discussions between professional instructors and five distinct learner profiles (child, teenager, college student, graduate student, and expert). We exclude these discussions' introductory and concluding portions, retaining only the dialogues between the instructor and the participants in each video.

The transcriptions contain around 570 minutes of 125 one-to-one conversations, revealing the depth and range of engagement between the 150 involved interlocutors. The transcriptions encompassed 102,656 words and consisted of 2,881 conversational turns, highlighting the dynamic back-and-forth of the educational discourse. Additionally, the gender distribution of the instructors in these videos was notably diverse, with 10 female instructors and 15 male. Table~\ref{tab:demography} shows the gender distribution among all participants at different levels. The average length of these videos is 23 minutes. 

\section{Conversational Analysis}
Conversation analysis (CA) explores how interactions are structurally organized among interlocutors~\cite{okeke2020learning}. For videos, it uses text-level metrics like the number of turns, words, and sentences for each utterance. We also calculate standard readability metrics like Flesch-Kincaid Readability Ease (FKRE) and Flesch-Kincaid Grade Level (FKGL) to explore the level of readability of transcription for each complexity level across utterances. FKRE and FLGL calculate with a weighted scoring formula that considers both sentence length and the number of syllables. The FKRE range from 0 to 100, with lower scores suggesting that the text is more challenging for readers to comprehend.~\cite{kincaid1975derivation}. For consistency and clarity, we labeled each instructor in the videos as 'INS.` followed by their respective sequence number of the episode. Our effort serves as the groundwork for teaching science concepts adapted to different levels of complexity, opening insights that can be used in how we impart knowledge and facilitate the development and integration of state-of-the-art computational solutions such as language models, and can be harnessed with precision and effectiveness. 


Table~\ref{tab:token-ratio} presents the ratio of words spoken by the instructor to those spoken by the learner for each video. Our analysis captures a notable trend: instructors dominate conversations when interacting with younger audiences, especially children and teenagers. In these interactions, 64\% of the instructors spoke at least three times more than their younger counterparts. However, when conversing with more knowledgeable groups such as graduate students, only 16\% of instructors spoke more than 3 times the learners. Intriguingly, at the expert level, the trend reversed, with 72\% of experts speaking more than the instructors. 
Furthermore, an observable trend in some videos, highlighted in Table~\ref{tab:token-ratio}, is the decreasing ratio of words spoken by instructors compared to learners as the audience's proficiency level rises from child to expert. This suggests that the instructors tend to speak less as the learner's expertise increases, allowing the more knowledgeable participants to contribute more to the conversation. It highlights the dynamic nature of instructor-learner interactions, adapting based on the audience's knowledge level.

Among all the videos, videos 8 and 25 tackle the topic of black holes. Some noteworthy trends emerge when examining the word count ratio of instructors to learners across different educational levels. For video 8, presumably a more basic explanation, the word count ratio is the highest for children at 3.69, indicating a more detailed or elaborative teaching approach. As the educational level progresses to teens and college students, the ratio drops, with the least for graduates at 1.05. In contrast, video 25 shows a different word count distribution. Here, the ratio is highest for teens at 5.34 and least for graduates at 0.95, suggesting that different instructors might have different teaching styles, even over the same topic.


\begin{table}[!ht]
    \centering
    \begin{tabularx}{\columnwidth}{l*{5}{c}} 
        \toprule
        \textbf{Video} & \textbf{child} & \textbf{teen} & \textbf{coll.} & \textbf{grad.} & \textbf{exp.} \\
        \midrule
        1 & 3.36 & 9.84 & 1.65 & 1.86 & 1.12 \\
        \rowcolor{Gray}2 & \multicolumn{1}{c}{\bfseries 11.19} & \multicolumn{1}{c}{\bfseries 7.65} & \multicolumn{1}{c}{\bfseries 3.36} & \multicolumn{1}{c}{\bfseries 2.24} & \multicolumn{1}{c}{\bfseries 0.8} \\

        3 & 2.83 & 12.05 & 5.25 & 1.15 & 1.02 \\
        4 & 4.85 & 5.23 & 3.78 & 0.49 & 0.54 \\
        5 & 1.87 & 5.73 & 2.16 & 0.93 & 0.55 \\
        6 & 2.89 & 3.52 & 1.93 & 2.63 & 0.49 \\
       \rowcolor{Gray} 7 & \multicolumn{1}{c}{\bfseries 9.31} & \multicolumn{1}{c}{\bfseries 7.46} & \multicolumn{1}{c}{\bfseries 2.50} & \multicolumn{1}{c}{\bfseries 1.36} & \multicolumn{1}{c}{\bfseries 0.94} \\
        8 & 3.69 & 1.25 & 1.35 & 1.05 & 1.32 \\
        9 & 1.09 & 4.83 & 8.04 & 1.18 & 0.72 \\
        10 & 2.61 & 4.57 & 3.97 & 3.37 & 1.45 \\
        11 & 5.57 & 2.91 & 0.44 & 2.04 & 0.38 \\
        12 & 1.08 & 1.25 & 4.35 & 0.72 & 0.58 \\
       \rowcolor{Gray} 13 & \multicolumn{1}{c}{\bfseries 6.25} & \multicolumn{1}{c}{\bfseries 1.93} & \multicolumn{1}{c}{\bfseries 1.13} & \multicolumn{1}{c}{\bfseries 0.48} & \multicolumn{1}{c}{\bfseries 0.96} \\
        14 & 3.02 & 1.82 & 2.68 & 0.34 & 0.86 \\
        15 & 2.29 & 1.66 & 3.68 & 1.56 & 0.83 \\
        \rowcolor{Gray} 16 & \multicolumn{1}{c}{\bfseries 6.91} & \multicolumn{1}{c}{\bfseries 4.16} & \multicolumn{1}{c}{\bfseries 0.91} & \multicolumn{1}{c}{\bfseries 0.61} & \multicolumn{1}{c}{\bfseries 0.35} \\
        17 & 0.15 & 0.29 & 3.93 & 3.33 & 0.51 \\
        18 & 4.6 & 1.99 & 7.51 & 0.58 & 0.29 \\
        19 & 3.03 & 1.19 & 0.26 & 4.20 & 1.10 \\
        20 & 7.05 & 4.35 & 5.07 & 2.20 & 0.78 \\
        \rowcolor{Gray} 21 & \multicolumn{1}{c}{\bfseries 4.98} & \multicolumn{1}{c}{\bfseries 3.80} & \multicolumn{1}{c}{\bfseries 3.34} & \multicolumn{1}{c}{\bfseries 1.34} & \multicolumn{1}{c}{\bfseries 0.92} \\
        22 & 6.81 & 7.63 & 3.60 & 5.32 & 0.94 \\
        \rowcolor{Gray} 23 & \multicolumn{1}{c}{\bfseries 6.83} & \multicolumn{1}{c}{\bfseries 4.82} & \multicolumn{1}{c}{\bfseries 3.90} & \multicolumn{1}{c}{\bfseries 1.17} & \multicolumn{1}{c}{\bfseries 0.49} \\
        \rowcolor{Gray} 24 & \multicolumn{1}{c}{\bfseries 11.43} & \multicolumn{1}{c}{\bfseries 5.17} & \multicolumn{1}{c}{\bfseries 2.90} & \multicolumn{1}{c}{\bfseries 1.53} & \multicolumn{1}{c}{\bfseries 1.50} \\
        25 & 2.19 & 5.34 & 3.11 & 0.95 & 1.96 \\
        \bottomrule
\end{tabularx}
\caption{Instructor to learner word count ratio: highlighted rows show decreasing ratio as audience proficiency increases.}
\label{tab:token-ratio}
\end{table}

\begin{table}
    \centering
    \begin{tabularx}{\columnwidth}{|l|X|X|X|X|X|}
        \hline
         & child & teen & coll. & grad. & exp. \\
        \hline
        INS.1  & 84.57 & 72.66 & 71.04 & 56.29 & 55.24 \\
        \hline
        INS.3  & 85.59 & 74.79 & 63.90 & 62.27 & 61.16 \\
        \hline
        INS.22  & 88.16 & 75.13 & 69.19 & 65.96 & 58.50 \\
        \hline
    \end{tabularx}
    \caption{Example FKRE metrics of instructor turns when teaching various learners.}
    \label{tab:FKRE}
\end{table}

Our study measures different readability scores associated with instructors and learners. The investigations show how instructors tailor their language complexity to suit different audiences. Notably, the FKRE of instructors, such as INS.1 and INS.3, demonstrated reduced value and increased complexity as the conversational level progressed from a child to an expert, as shown in Table~\ref{tab:FKRE}. For example, in Video 1, the FKRE scores revealed a pattern where the child had a higher readability score of 84.57, gradually decreasing to 55.24 for the expert. 


\subsection{Presentation styles}
The presentation of scientific concepts across these videos varies in depth, complexity, and technical jargon. For children, educators often start with questions about whether they know the concept. For instance, INS.9 asked a child, ``Do you know what gravity is?''. Similarly, INS.10 initiated a conversation with, ``Have you heard the word `physics' before? Do you know what that means?''. Simple language, visual content, and real-world analogies are commonly used. The emphasis is on making the concept engaging, relatable, and easily digestible by using storytelling and a playful approach to explain it. For example, INS.1 uses a conversation with her son to explain the concept of Nuclear Fusion to the child learner. At the teen level, the depth of content expands. Analogies are more sophisticated, and some basic technical terms have been introduced. Instructors attempt to connect the new topic to the teens' school knowledge with questions such as, ``Have you played with magnets in school?'' posed by INS.6, or ``So, have you had much exposure to lasers yet at school?'' asked by INS.20. The concept presentations to the college and graduate schools are more specialized. Educators assume a foundational understanding of the subject and thus delve into more detailed information. Technical jargon is extensive, and the content might often be segmented into various sub-topics on the topic. For expert-level explanations, educators mostly ask them to explain highly specialized concepts and the cutting-edge aspects of the subject. Discussions often involve current research and future implications. Rather than basic explanations, the content is discussion-driven, analysing the topic's nuances and frontiers. 

The use of analogies and metaphors varies across different audiences and levels. For Example, INS.2 explains the concept of time to a child by comparing the relation between aging and the number of Earth's revolutions around the sun, or INS.9 explains gravity as: ``It's something that, so, right now, we would be floating if there was no gravity, but since there's the gravity we're sitting right down on these chairs.''


College and graduate-level analogies aim to bridge the gap between fundamental concepts and advanced theories. When INS.2 explains the time to the graduate learner, it could be likened to a billiard ball metaphor, as ``it goes into a wormhole, comes out and hits the ball going into the hole. And in that way, if it could knock it off course, we seem to be in some logical paradox''. 

Instructors use various strategies to adapt their teaching to specific audiences. While no single strategy fits all, we observe many common behaviors across these conversations. They begin by assessing the prior knowledge of their audience to determine the starting point. Instructors use simplified content using basic vocabulary, some visuals,  hands-on activities, and real-world examples to teach scientific concepts to younger learners. College and graduate students are assumed to have a foundation in the subject. They can understand abstract notions and are ready for more in-depth discussions and evaluations. Expert learners focus on the latest research, sharing ideas, unresolved questions in the field, and collaborative projects. Discussions can be highly technical and specialized. Instructors frequently tap into the curiosity of less knowledgeable learners, such as children or teens, by crafting examples and activities that enable them to explore scientific concepts independently. After introducing a concept, they collect feedback with questions from these learners to assess their understanding. For instance, INS.7 might ask, ``What have you learned about black holes?''.

\section{Conclusion}
We present a 5-Levels dataset and its analysis containing transcriptions of discussions featured on teaching scientific concepts with various difficulty levels. It serves as a resource for researching educational interactions within the context of adapting scientific topics to varying complexity levels, a new domain that has been neglected in prior research. We use CA techniques to analyse different turns of instructors and learners. Educators use various strategies to tailor their teaching styles to these different complexity levels. Analogies and metaphors are valuable tools for conveying complex ideas, becoming more abstract and technical as the audience's knowledge deepens. This work addresses the absence of suitable datasets for more computational approaches and training language models to aid adaptive teaching practices.

\section*{Ethical Considerations}
Age is a protected category. However, it is inextricably linked to the study of education in most cultures. In our study, we use data from the available series of videos about teaching scientific concepts to different levels of difficulty on the WIRED website, and evaluate the transcripts via automated text analysis metrics. As the data is published presumably with the consent of the subject, there is minimal risk of abuse and no concerns for the welfare of human subjects in our analysis.


\nocite{*}
\section{References}\label{sec:reference}
\bibliographystyle{lrec-coling2024-natbib}
\bibliography{lrec-coling2024-example}

\end{document}